\documentclass[10pt,twocolumn,letterpaper]{article}
\usepackage{arxiv}
\usepackage{graphicx}
\usepackage{placeins}
\usepackage{amsmath}
\usepackage{amssymb}
\usepackage{booktabs}
\usepackage{array}
\usepackage{bm}
\usepackage[pagebackref,breaklinks,colorlinks]{hyperref}
\usepackage[capitalize]{cleveref}
\usepackage[title]{appendix}

\newcolumntype{x}[1]{>{\centering\arraybackslash\hspace{0pt}}p{#1}}

\crefname{section}{Sec.}{Secs.}
\Crefname{section}{Section}{Sections}
\Crefname{table}{Table}{Tables}
\crefname{table}{Tab.}{Tabs.}
\Crefname{equation}{Equation}{Equations}
\crefname{equation}{eq.}{eqs.}

\newcommand{\bx}{\bm{x}}
\newcommand{\bv}{\bm{v}}

\makeatletter
\renewcommand{\paragraph}{%
  \@startsection{paragraph}{4}%
  {\z@}{0.25em}{-1em}%
  {\normalfont\normalsize\bfseries}%
}
\makeatother

\begin{document}

\title{Real-time volumetric rendering of dynamic humans}

\author{Ignacio Rocco\quad\quad Iurii Makarov\quad\quad Filippos Kokkinos\quad\quad David Novotny\\ Benjamin Graham\quad\quad Natalia Neverova\quad\quad Andrea Vedaldi\\\,\\Meta AI}
\maketitle

\begin{abstract}
We present a method for fast 3D reconstruction and real-time rendering of dynamic humans from monocular videos with accompanying parametric body fits.
Our method can reconstruct a dynamic human in less than 3h using a single GPU, compared to recent state-of-the-art alternatives that take up to 72h.
These speedups are obtained by using a lightweight deformation model solely based on linear blend skinning, and an efficient factorized volumetric representation for modeling the shape and color of the person in canonical pose.
Moreover, we propose a novel local ray marching rendering which, by exploiting standard GPU hardware and without any baking or conversion of the radiance field, allows visualizing the neural human on a mobile VR device at 40 frames per second with minimal loss of visual quality.
Our experimental evaluation shows superior or competitive results with state-of-the art methods while obtaining large training speedup, using a simple model, and achieving real-time rendering. 
\end{abstract}

\section{Introduction}%
\label{sec:intro}

\begin{figure}
\centering
\includegraphics[width=\columnwidth]{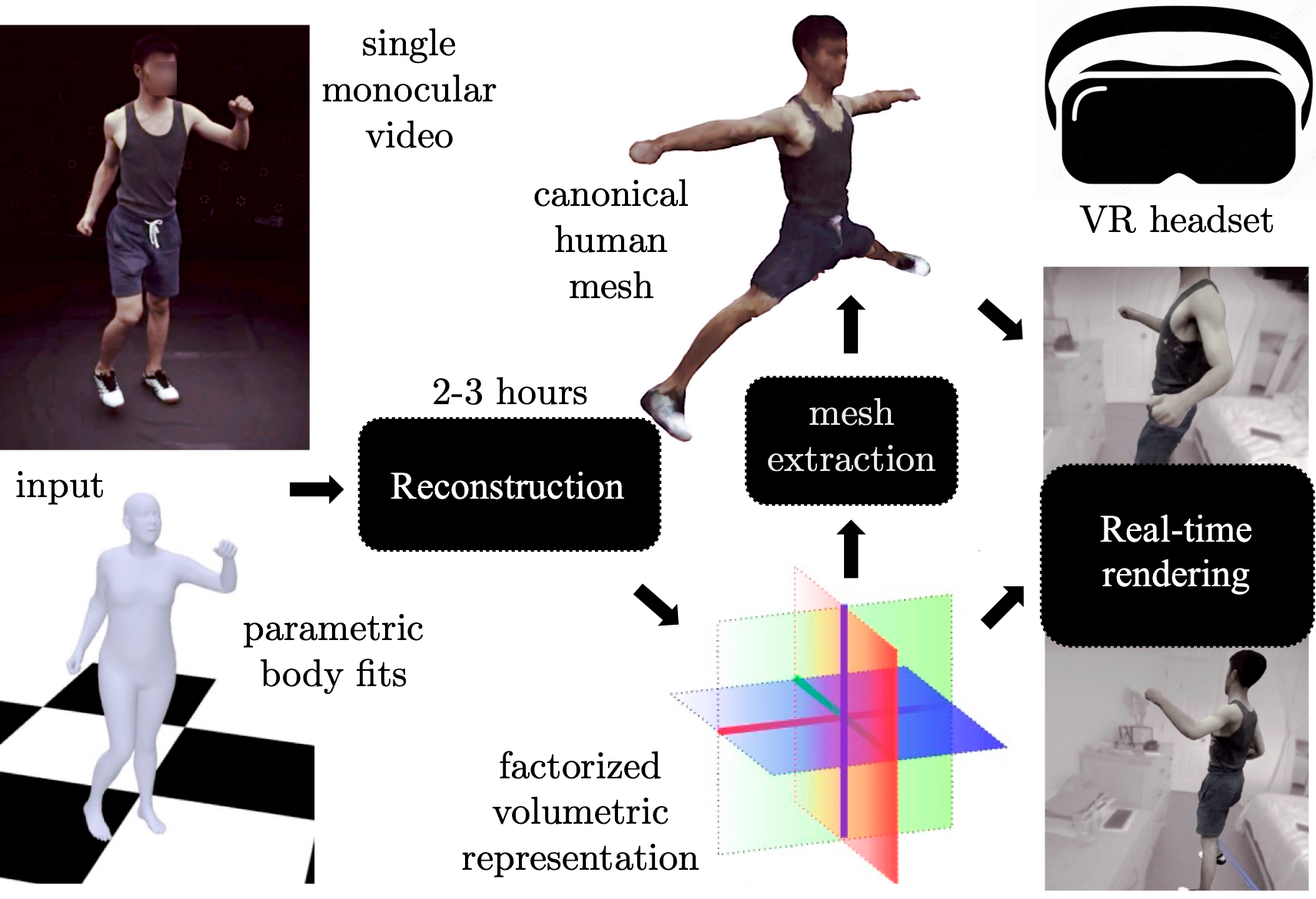}
\caption{
Given a monocular video with a SMPL parametric body fit, our method quickly reconstructs a 3D model of the person in canonical pose using a factorized radiance field representation~\cite{tensorf}. Then, it extracts a tight rigged mesh and uses a new local ray-marching render which, without baking or otherwise converting the radiance field, can render the dynamic human on a consumer VR device at 40FPS. Videos and demos are available at \url{https://real-time-humans.github.io/}.\label{fig:teaser}}
\end{figure}

Emerging technologies such as virtual and mixed reality make photorealistic 3D reconstruction increasingly relevant and impactful.
We can envisage a future in which, instead of taking pictures or movies with smartphone cameras, anyone will be able to capture and experience full $360^\circ$ holograms with equal quality, simplicity, and generality.
However, despite recent progress in methods such as neural rendering, we remain quite far from this objective.

Many neural rendering approaches assume static scenes and the availability of dozens if not hundreds of input views for each reconstructed scene~\cite{nerf,mipnerf,mipnerf360,FVS,SVS,neus}.
Many dynamic extensions of neural rendering~\cite{nerfies,hypernerf,dnerf,nsff,nguyen2021human} handle small deformations.
The most interesting content, however, is highly dynamic and often involves people.
This has motivated the development of specialised models that can account for the highly-deformable structure of such objects, but most of these still assume that scenes are captured from multiple cameras~\cite{neuralactor,neuralbody,tava, humannerf-shanghai}, which is incompatible with consumer applications.
When data is captured by a smartphone or pair of AR glasses, only a single slowly-varying viewpoint is available for 3D reconstruction.

A few methods such as \mbox{HumanNeRF-UW}~\cite{humannerf-uw}, \mbox{A-NeRF}~\cite{anerf} and NeuMan~\cite{neuman} can obtain photorealistic reconstructions of articulated humans from such monocular videos.
These methods use articulated human models such as SMPL~\cite{smpl} to define a \emph{scaffold} to model the object deformation, using a pre-trained predictor to obtain an initial estimate of the 3D body shape and pose.
Then, they capture the shape of the subject over time as a deformation of a \emph{canonical} radiance field, which is time-invariant.
Unfortunately, jointly learning a radiance and deformation fields with a neural network requires several days of optimization to reconstruct a single monocular video.
Furthermore, 3D video playback using the standard emission-absorption method is also slow due to the necessity of evaluating the neural network hundreds of times for each rendered pixel.

In this paper, we reconsider these architectures and we improve them significantly in terms of training and rendering speed, achieving fast training and real-time rendering.
This is achieved by three design decisions, discussed next.

First, instead of representing the radiance field with a neural network, we use a tensor-based decomposition of the field, inspired by~\cite{tensorf,pifu,eggan}.
This representation leads to faster convergence than when using an MLP\@.

Second, we directly leverage linear blend skinning (LBS) in SMPL to  model motion during training and rendering.
This differ from previous approaches that trained a module to invert SMPL~\cite{humannerf-uw,tava} or learned a 3D scene flow field using a separate MLP~\cite{humannerf-uw,neuralactor,neuman}, which is costly.
Instead, we map each 3D point to canonical space \emph{solely} by approximating the inverse LBS function using the inverse transformation of the closest SMPL face.
This approximation is good where it matters, namely when the queried 3D point is already in the vicinity of the body surface.

Third, we propose to extract a personalized human mesh template from the reconstructed human in the factorized radiance field.
During training, the exact shape of the person is not known, and the continuous volumetric representation allows to learn it though back-propagation.
However, neural rendering requires sampling the full estimated object bounding box, which is too slow for real-time rendering.
After the training is completed, we thus extract a mesh that is close to the surface of the reconstructed person.
This mesh can then be used as a tight scaffold to implement neural rendering efficiently on device.
Specifically, we describe a new local ray marching algorithm that carries out, using a custom shader, ray marching and emission absorption on a short distance from the mesh triangles as these are processed by the GPU rasteriser.
Compared to other recent hardware-accelerated neural rendering techniques, our technique avoids baking, \ie, sampling the radiance field values and storing them in a different specialized data structure, which is usually expensive and lossy.
Instead, it displays the \emph{unmodified} radiance field directly.

To summarize, our \textbf{contributions} are the following:
\textbf{1.} We show that adopting a factorized radiance field representation and a simple LBS-based deformation model allows for fast reconstruction ($24$ times faster than HumanNeRF-UW) with comparable or better rendering quality on the ZJU-Mocap~\cite{neuralbody} scenes.
\textbf{2.} We show that it is possible to extract a canonical mesh from the learned radiance field, together with a rig derived from the LBS model in SMPL, which approximates well each individual.
\textbf{3.} We show that, given these design choices, it is possible to use GPU hardware to renderer the deformable radiance field model preserving quality while achieving real-time playback (at 40 FPS on the mobile GPU of a consumer VR device), which is three orders of magnitude faster than the HumanNeRF-UW approach (0.05 FPS)\@.

\section{Related work}%
\label{sec:related_work}

\paragraph{Reconstructing humans from single images.}

Performing 3D reconstruction of humans from single images is ill-posed and requires leveraging prior information on human geometry and poses.
Several methods~\cite{smplify,romp,spin} use the SMPL model~\cite{smpl} as prior, but they result in only approximate reconstructions which are insufficient for VR/AR applications.
PIFu~\cite{pifu,pifuhd} directly predicts a volumetric representation of a human from a single RGB image using an MLP\@, but generalizes poorly beyond the training data distribution.
ARCH~\cite{arch,archpp} combines parametric models and implicit function representations to obtain higher fidelity reconstructions, which are also rigged and animatable, but achieve limited overall quality due to the fact that they use a \emph{single} image and cannot take advantage of video data.

\paragraph{Reconstructing humans from multiple videos.}

In order to obtain higher quality reconstructions, several methods~\cite{neuralactor, neuralbody, tava, humannerf-shanghai} leverage multiple synchronized videos.
Neural Actor~\cite{neuralactor} learns neural radiance fields~\cite{nerf} to model a human in canonical pose, and uses the inverse linear blend skinning (LBS) transform from SMPL to map between posed and canonical spaces.
TAVA~\cite{tava} uses a similar approach, but proposes to learn the inverse LBS weights and use Mip-NeRF~\cite{mipnerf} instead of vanilla NeRF\@.
Neural Body~\cite{neuralbody} attaches neural codes to the SMPL vertices, poses them, and converts them to a full volumetric representation of the radiance field using a 3D sparse CNN~\cite{sparseconvnet, minkowskiengine}.
HumanNeRF-ShanghaiTech~\cite{humannerf-shanghai} combines NeRF and the inverse LBS mapping (as in Neural Actor), but conditions the NeRF model on features sampled from neighbouring input views similar to IBRNet~\cite{ibrnet} for rigid scenes.
However, these methods require multiple views, and are thus inapplicable when only a monocular sensor is available.

\paragraph{Reconstructing humans from monocular videos.}

Recent methods considered 3D human reconstruction from \emph{monocular videos}.
Vid2Actor~\cite{vid2actor} extracts from SMPL a motion basis, uses it for inverse LBS, and trains two 3D CNNs to regress the inverse LBS weights in posed space and the density and color of the canonical human.
HumanNeRF-UW~\cite{humannerf-uw} extends Vid2Actor by replacing their voxel-grid representation with a coordinate MLP and further refine deformations using SMPL pose refinement and a non-rigid scene-flow MLP\@.
A-NeRF~\cite{anerf} computes codes of the posed 3D points that are relative to the SMPL skeleton and use those to construct a pose-dependent NeRF model (somewhat analogously to the approach used in Neural Body).
NeuMan~\cite{neuman} decomposes the scene into a static background component and the foreground dynamic human, and learns a different NeRF model for each component separately.
They use inverse LBS for warping points from posed to canonical space and a standard NeRF MLP for modelling the canonical human.

While powerful, these models are typically very slow to train (sometimes in the order of days on a single GPU) and to render, making them unsuitable for playback in mobile devices such as in VR headsets.

\paragraph{Real-time neural and volumetric rendering.}

Most of the works discussed above represent radiance fields using MLPs and use raymarching for rendering.
Hence, the MLPs need to be evaluated hundreds of time for each pixel, which requires seconds for each rendered image.
For real-time rendering, several methods have proposed to ``bake'' or cache the NeRF MLP~\cite{snerg, plenoctrees, fastnerf}.
However, such representations are memory intensive, do not support dynamic content, and require a high-end GPU for achieving real-time rendering.

Closer to ours, MobileNeRF~\cite{mobilenerf} leverages the fast triangle rasterization hardware in modern GPUs for real-time rendering of NeRF, but with substantial differences.
They also need to ``bake'' the radiance field into an especially-crafted mesh to avoid emission-absorption rendering.
Because they change the rendering model, they require a conversion step that involves further learning and approximations, which also does not support dynamic content.
In contrast, our technique renders in real time the \emph{original} dynamic radiance field, accelerating emission absorption.
Because we load the original model into the GPU, we do not require conversion, and we can preserve the shape and appearance details in the source radiance field with only minor approximations.
As far as we are aware of, our method is the first one to be able to render dynamic radiance fields in real-time on mobile hardware without resorting to baking.

\begin{figure*}
\centering
\includegraphics[width=\textwidth]{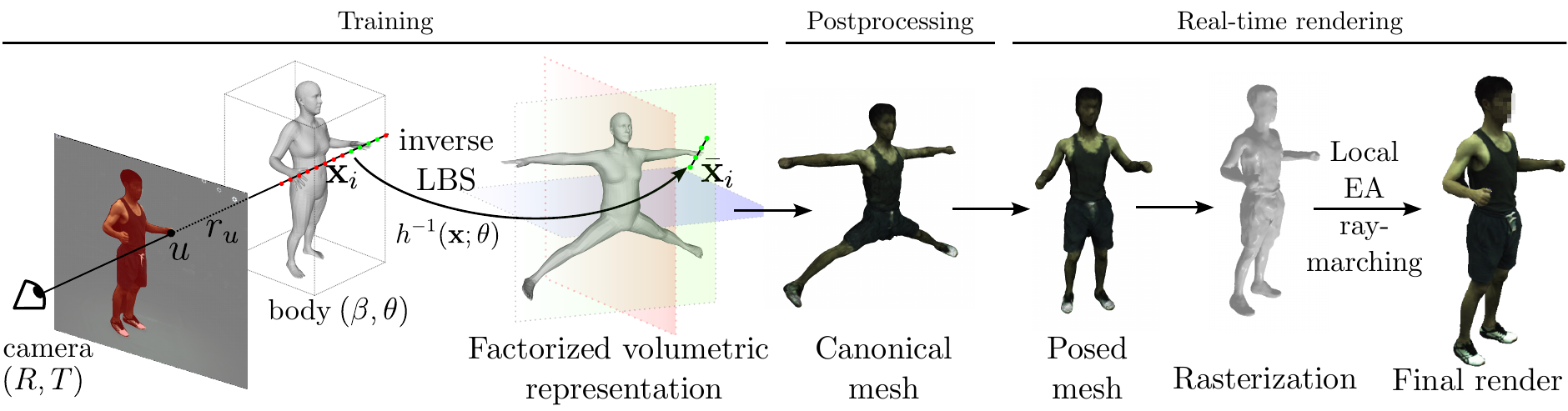}
\caption{\textbf{Overview of the proposed method.}
During training, we shoot rays from the training cameras onto the scene and sample points withing the bounding box of the parametric human body mesh.
Points which are close to the mesh are warped to canonical space via inverse linear-blend-skinning (LBS), where the factorized volumetric representation is sampled for density and color, and whose values are used to perform raymarching.
After training, a canonical mesh is extracted from the learnt factorized volumetric representation. For real-time rendering, the canonical mesh is posed, rasterized, and used to guide the inverse skinning and local raymarching.}%
\label{fig:model}
\end{figure*}

\section{Fast reconstruction of dynamic humans}%
\label{sec:proposed_method}

Our method leverages recent progress in neural fields and monocular 3D reconstruction in order to allow for fast 3D reconstruction of humans from monocular videos and corresponding parametric SMPL fits.
During reconstruction, our method employs the following modules:
(i) a factorized volumetric representation that learns the shape and color of the human in canonical pose,
(ii) a deformation module which performs inverse linear blend skinning to map 3D points from posed space to canonical space.
In addition, we propose
(iii) a rasterization-based method for rendering the trained model in real-time.
Our full model is illustrated in \cref{fig:model}.
Each of these components is presented next in more detail.

\subsection{Factorized volumetric radiance fields}%
\label{sec:radiance_fields}

In this work, we leverage recent progress on neural radiance fields~\cite{nerf,tensorf} as a continuous and differentiable representation for modeling shape and color. Next, we review these models and present the factorized formulation that we adopt in this work.  

\paragraph{Radiance fields.}

An image $I : \Omega \rightarrow \mathbb{R}^3$ is a map from pixels $u\in\Omega = [0,\dots,W] \times [0,\dots,H]$ to corresponding colors $I(u) \in \mathbb{R}^3$.
The camera projection function $\pi : \mathbb{R}^3 \rightarrow \mathbb{R}^2$ is a map from 3D points $\bx \in \mathbb{R}^3$ in the world space to corresponding pixels (image points) $u \in \mathbb{R}^2$.
A \emph{radiance field} is a pair of functions mapping each 3D point $\bx$ to a corresponding density $\sigma(\bx) \in \mathbb{R}_+$ and corresponding color $c(\bx) \in \mathbb{R}^3$.

The color of a pixel $u$ is obtained by \emph{marching} along the ray $r_u$ originating from the camera center and propagating in the direction of pixel $u$ (interpreted as a 3D point).
Let $(\bx_i)_0^{N-1}$ be a sequence of $N$ samples along ray $r_u$ separated by steps $\Delta$.
The color of pixel $u$ is extracted from the radiance field via the following emission-absorption rendering equation:
\begin{equation}\label{eq:rend}
  I(u)
  =
  \sum_{i=0}^{N-1}
  (T_i - T_{i+1}) c(\bx_i),
  ~~~
  T_i = e^{-\Delta \sum_{j=0}^{i-1} \sigma(\bx_i)}.
\end{equation}
where $T_i$ is the accumulated transmittance up to $\bx_i$, namely the probability that a photon is transmitted from point $\bx_i$ back to the camera center without being absorbed.

\paragraph{Deformable radiance fields.}

The radiance field of~\cref{eq:rend} could be extended to represent the different poses of an articulated object by adding the object's pose parameters $\theta$ as an additional parameter of the density $\sigma(\bx;\theta)$ and color $c(\bx;\theta)$ functions.
Modelling such functions directly, however, would be statistically inefficient because it would not account for the fact that the different poses are not arbitrary, but related by geometric deformations.

Therefore, in this work we adopt more efficient approach with  considers \emph{canonical} density $\bar \sigma(\bar{\bx})$ and color $\bar c(\bar{\bx})$ fields that are pose-invariant and, separately, a \emph{posing} function $\bx = h(\bar{\bx}; \theta)$ mapping points $\bar{\bx}$ from the canonical space to their posed locations $\bx$.
The pose-dependent fields are the composition of the pose-invariant fields and of the \emph{inverse} of the posing function:
\begin{equation}\label{eq:def-rf}
  \sigma(\bx;\theta) = \bar \sigma(h^{-1}(\bx;\theta)),
  ~~~
  c(\bx;\theta) =  \bar c(h^{-1}(\bx;\theta)).
\end{equation}

\paragraph{Tensorial fields.}

We leverage the recent factorized volumetric neural field formulation from TensoRF~\cite{tensorf} to represent the shape and color of the \emph{canonical} human.

In order to model the field $(\sigma,c)$, we do not use the standard approach of adopting an MLP and positional encoding~\cite{nerf}, but use instead the TensoRF parameterization.
Specifically, we consider a \emph{voxel grid} of resolution $D\times H\times W$ and further decompose it as the sum of three matrix-vector products:
\begin{equation}\label{eq:tensorf-sigma}
\bar \sigma(\bar \bx)
\!=\!
\rho \left(
  \sum_{r=1}^{R_\sigma}\!
  M^{YX}_{r,\bar y, \bar x} v^{Z}_{r,\bar z}\!+\!  M^{YZ}_{r,\bar y, \bar z} v^{X}_{r,\bar x}\!+\!
  M^{XZ}_{r,\bar x, \bar z} v^{Y}_{r,\bar y}\!
\right)\!.
\end{equation}
Here $\bar{\bx} = (\bar x, \bar y, \bar z)$, $r$ is the channel index, and the sub-indices indicate the tensor sampling position.\footnote{In order to map coordinates $\bar{\bx} = (\bar x, \bar y, \bar z)$ defined in canonical space to indices in the respective tensors and matrices above, we use bilinear interpolation after remapping the nominal bounding box of the object to the corresponding grid dimensions.} In addition, $M^{YX}$ is a $R_\sigma \times H\times W$ tensor, $v^{Z}$ a ${R_\sigma}\times D$ matrix, and the other terms follow a similar pattern, so in total there are $R_\sigma(HW+HD+WD+H+W+D)$ parameters only, far less than $HWD$ as long as the number of components $R_\sigma \ll (HWD)^\frac{1}{3}$.
The activation function $\rho(a) = \log(1 + \exp(G a))$ is the \textit{softplus} operator with a fixed gain $G \approx 1/\Delta$, where $\Delta$ is the step between ray samples.

The color field $\bar c$ is defined in a similar manner, for each of the three RGB components, and uses the sigmoid as activation function.

The adoption of these factorized representation not only allows for faster training compared to the MLP-based models, but is \emph{particularly suitable} for fast real-time rendering by storing the different tensor factors as textures that are naturally interpolated by the  graphic shaders of GPUs (cf.~Sec.~\ref{sec:rt}).

\subsection{Skinning-based deformation module\label{sec:deformation}}

Several previous works use MLP-based deformation models~\cite{nerfies, hypernerf, humannerf-uw, neuralactor,neuman}, alone, or in conjunction with articulated deformations~\cite{neuralactor,humannerf-uw,neuman}.
Because these models are slow to evaluate, we propose to use only a linear blend skinning (LBS) articulated deformation model, and show that this simple deformation model can still achieve competitive results while being much simpler and allowing for real-time rendering.
We next review LBS and present our proposed approach.

\paragraph{Posing via blend skinning.}

Given that humans are articulated objects, we leverage the parametric SMPL model~\cite{smpl} and the linear blend skinning formulation which allows to easily build the posing function $h$, given the SMPL template mesh model and the pose parameters.
In particular, the SMPL model provides a template mesh $\mathcal{M}=(\mathcal{V},\mathcal{F})$ with points ${\bar{\bv}} \in \mathcal{V} \subset \mathbb{R}^3$ and triangular faces $f \in \mathcal{F}$.
In addition, the template mesh is associated with a \emph{skeleton}, which is a collection of bones $b\in \{ 1,\dots,B \}$ whose angles define the pose parameters.
Each template vertex $\bar{\bv}$ is~\emph{softly} attached to the different bones by its \emph{skinning weights} $w(\bar{\bv})\in\mathbb{R}_+^B$.
Then, the posed vertices can be obtained by an affine map
$
\bv = h(\bar{\bv}; \theta) = A(\bar{\bv} ; \theta) \bar{\bv}\
$,
where:
\begin{equation}\label{eq:fwd}
  A(\bar{\bv}; \theta)
=
\sum_{b=1}^B
w_b(\bar{\bv})
A_b(\theta),
\end{equation}
and $A_b(\theta)$ is the affine transform associated to bone $b$ for the pose $\theta$.
More details are provided in Appendix~\ref{appendix:lbs}.

\paragraph{Volumetric and inverse blend skinning.}

Skinned models like SMPL only define the skinning weights $w(\bar {\bv})$ for the vertices of the template mesh, which approximates the real object surface. 
However, because the radiance field is a volumetric representation, we need transformations to be defined \emph{around} the object surface, and not just \emph{on} the surface, and thus the skinning weights need to be extended to nearby 3D points.
Furthermore, in order to perform ray marching and compute the pose-invariant fields from \cref{eq:def-rf}, we need to map points $\bx$ from the posed space \emph{back} to canonical space, and therefore knowledge of the posing function $h(\bar{\bx}; \theta)$ is not enough; instead, we require its \emph{inverse}:
\begin{equation}
  \bar{\bx} \!=\! h^{-1}(\bx;\theta)
  \!=\!
  A(\bar{\bx} ; \theta)^{-1} \!\bx
  \!=\!\!
\left[
  \sum_{b=1}^B
w_b(\bar{\bx})~
\!A_b(\theta)
\right]^{-1}%
\!\!\!\!\!\!\!\cdot \bx.%
\label{eq:inv_lbs}
\end{equation}
Because $\bar{\bx}$ appears on the l.h.s.~and r.h.s.~of this expression, this defines $\bar{\bx}$ as the solution to an equation which cannot be solved in closed form.
Prior works~\cite{deng20nasa,chen21snarf:} addressed these issues by \emph{learning} the extended skinning weights and/or the inverse posing function, but these approaches increase the complexity of the model and therefore the training and rendering time.

\begin{figure}
\includegraphics*[width=0.9\columnwidth]{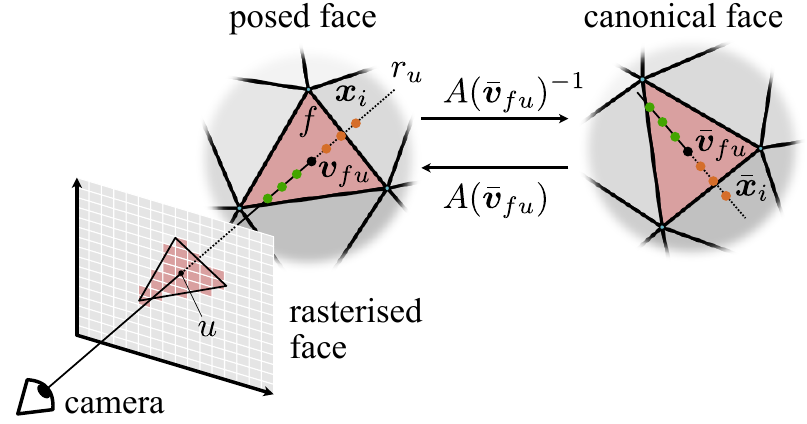}
\caption{\textbf{Real-time radiance fields via rasterization.} The posed face $f$ is used to determine the portion of the radiance field that is likely to affect given pixel $u$ and to map computations to the rasterization unit of the GPU.}\label{fig:rt}
\end{figure}

In contrast, we employ the approach introduced in~\cite{neuralactor}, where the inverse transformation of a point $\bx$ is approximated to that of the closest mesh vertex $\bar{\bv}_{i^*}$:
\begin{align}
    \bar{\bx} = h^{-1}(\bx;\theta) \approx A(\bar{\bv}_{i^*};\theta)^{-1} \bx,\nonumber\\
    \text{s.t.}~i^*=\operatornamewithlimits{argmin}_{i} \| \bx - \bv_i \|_2.
\end{align}
This approximation is only valid locally, and therefore only applied to ray points $\bx$ where the distance to the closest vertex is below a threshold $\tau$.
Otherwise, points are discarded and not used for raymarching, which is equivalent to setting
$
\sigma(\bx;\theta)
= \mathbf{1}_{[d(\bx, {\bv}_{i^*}) \leq \tau]} ~ \bar \sigma(h^{-1}(\bx;\theta)).
$

Contrary to prior work, we do not combine this deformation with any additional trainable model, in order to allow for real-time evaluation of the deformation model in mobile hardware. Our experimental results show that our method performs similarly to prior state-of-the-art despite of this simplified deformation model.
\FloatBarrier

\section{Real-time dynamic radiance fields}%
\label{sec:rt}

The factorized radiance field and deformation model presented in Sec.~\ref{sec:radiance_fields} and Sec.~\ref{sec:deformation} were chosen specifically to allow for real-time rendering, through a customized \emph{GPU shader} program that implements an efficient local emission-absortion raymarching.
Our approach \emph{avoids baking or conversion steps} and allows to replay the original radiance field with only minor approximations. 

\paragraph{Customized human mesh.}

Our proposed local raymarching is guided by an initial rasterization step of the posed mesh. In order to handle occlusions and object boundaries properly, we replace the SMPL mesh with one that more accurately approximates the body shape at rest, and transfer to the SMPL joints and blendshapes so that it is fully-rigged and amenable for posing.

We extract this mesh from the radiance field by:
(i) rendering frames with masks and depth-maps from the canonical model,
(ii) unprojecting these depth maps to form a dense point cloud and,
(iii) converting this point cloud onto a mesh. An example of the extracted mesh is shown on \cref{fig:model}.
Please refer to the Appendix~\ref{appendix:mesh} for further details.

\paragraph{Rasterization of the posed mesh.}

Without any guidance, volumetric rendering using~\cref{eq:rend} requries to sample hundreds of ray points for each generated pixel.
For an opaque object, this is extermely inefficient, as only a tiny fraction of those samples land close enough to the surface to make a significant contribution to the final color.
Instead, we resort to rasterization of the customized posed mesh as an initial step of our rendering pipeline, which provides guidance for performing a local volumetric rendering. 

In this case, the rasterizer iterates over the triangles $f\in\mathcal{F}$ of the mesh, quickly finding the pixels $u$ that are contained in them (\cref{fig:rt}).
The rasterizer also quickly computes the \emph{barycentric coordinates} $\alpha_i$ of the 3D face point $\bv_{fu} \in f$ that projects onto pixel $u$, defined as:
\begin{equation}\label{eq:bary}
  \bv_{fu} = \sum_{i=1}^3\alpha_i \bv_{fi},
  ~~~
  \alpha_i \geq 0,
  ~~~
  \sum_{i=1}^3 \alpha_i = 1,
\end{equation}
where $\bv_{fi}$ are the three mesh vertices.
Then, the color $I(u)$ of the pixel is obtained by computing the color $c(\bv_{fu})$ of point $\bv_{fu}$.
This calculation is done in parallel for all pixels using a \emph{shader program} running in the GPU. 

\paragraph{Local emission-absorption raymarching.} 
Instead of computing the color $c(\bv_{fu})$ by using an UV-texture mapping, as it is usually done for coloring meshes, we compute $c(\bv_{fu})$ by performing a local emission-absorption raymarching in the vicinity of point $\bv_{fu}$, as illustrated in \cref{fig:rt}. 

In practice, we first map the point $\bv_{fu}$ from the posed mesh to its counterpart in canonical space $\bar{\bv}_{fu}$ by applying the transformation $A(\bar{\bv}_{fu})^{-1}$. For this, $A(\bar{\bv}_{fu})^{-1}$ is efficiently and automatically approximated by the \emph{vertex shader} through barycentric interpolation, by storing the inverse transformation $A(\bar{\bv})^{-1}$ of each vertex $\bar{\bv}\in \mathcal{V}$ as a vertex property.

Once $\bar{\bv}_{fu}$ is computed, we can consider a small number of samples $\bar{\bx}_i$ (as illustrated in the right side of \cref{fig:rt}) to perform local emission-absorption raymarching. For this, the evaluation of the factors $M$ and $v$ of the density and color for fields can be done very efficiently on the GPU by mapping these 2D and 1D tensors to 2D and 1D \emph{textures maps} and using the native GPU texture sampling functions. 

Once the densities and colors are computed for all samples $\bar{\bx}_i$ in the local ray segment, \cref{eq:rend} is used to obtain the final pixel color $I(u)$.

\begin{figure*}
\centering
{\small
\setlength{\tabcolsep}{0pt}
\begin{tabular}{c}
 \includegraphics[width=\textwidth]{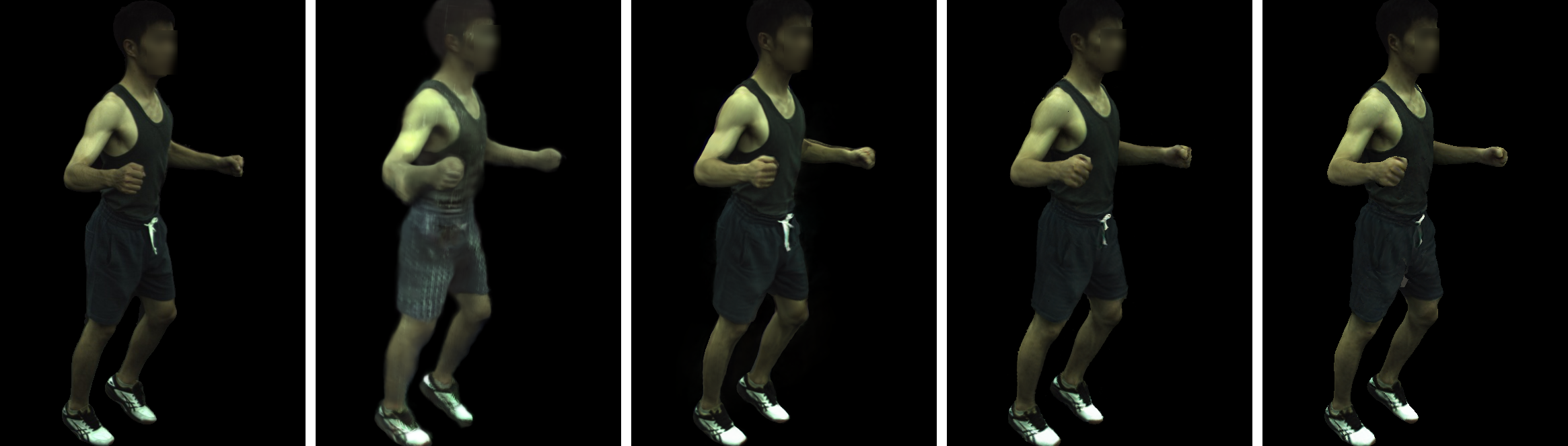} \\
 \includegraphics[width=\textwidth]{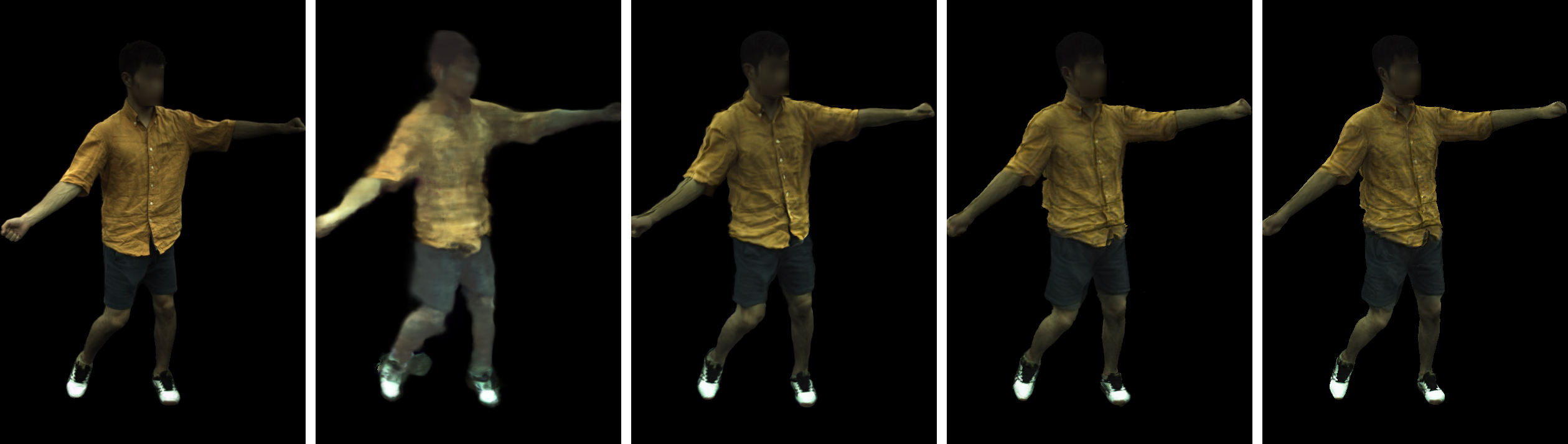} \\
 \includegraphics[width=\textwidth]{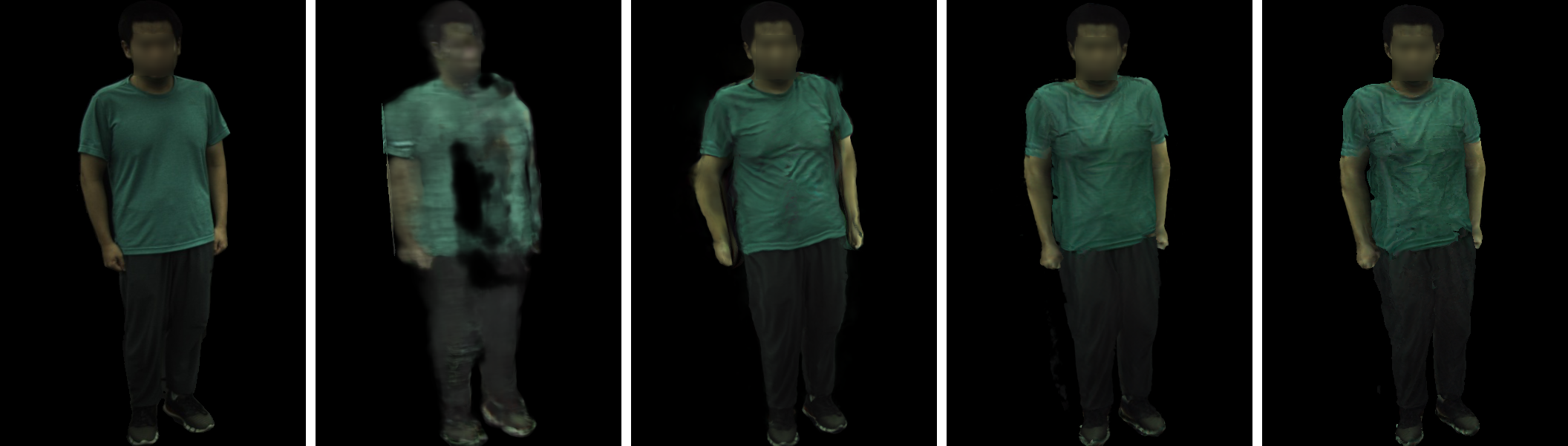} \\
 \includegraphics[width=\textwidth]{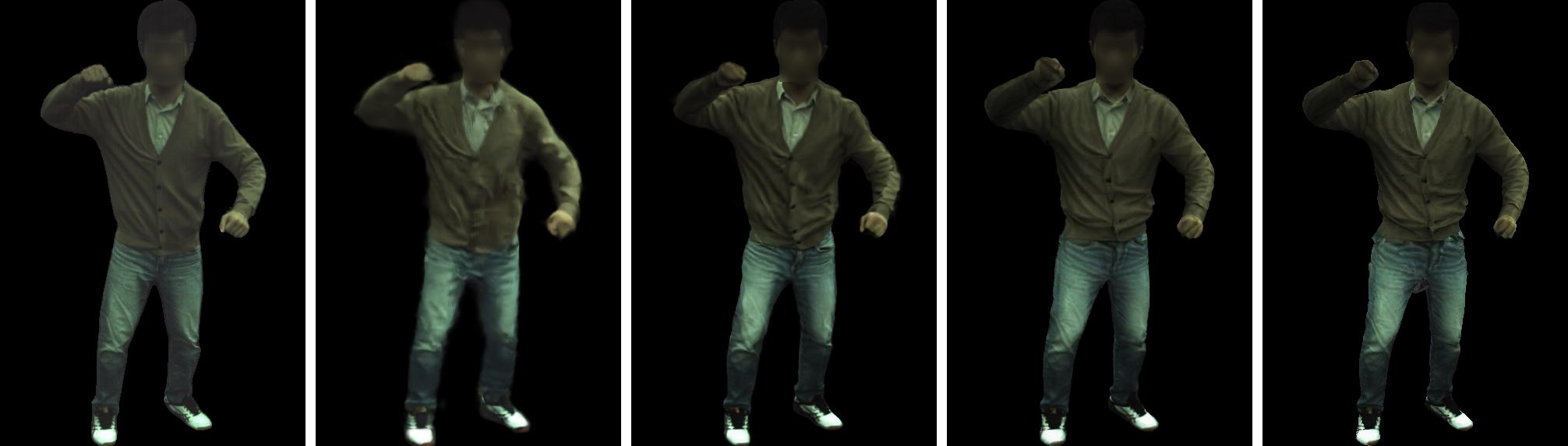} \\
\end{tabular}
{\setlength{\tabcolsep}{0pt}
\begin{tabular}{p{3.5cm}p{3.5cm}p{3.5cm}p{3.5cm}p{3.5cm}}
\centering (a) Ground-truth & 
\centering (b) NeuralBody~\cite{neuralbody} &
\centering (c) HumanNeRF-UW~\cite{humannerf-uw}  & 
\centering (d) Ours - reconstruction & 
\centering (e) Ours - real-time
\end{tabular}
}
}
\caption{\textbf{Qualitative results on ZJU-Mocap.}
We show a qualitative comparison between our proposed method and other state-of-the-art methods on test images.
Our method produces sharp images, perceptually similar to those of HumanNeRF-UW, despite presenting a simpler deformation model. Faces have been blurred for privacy reasons.}%
\label{fig:zju_qual}
\end{figure*}

\begin{table*}
\resizebox{\textwidth}{!}{
{\footnotesize
\setlength{\tabcolsep}{2pt}
\begin{tabular}{@{}lcccccccccccccccccc@{}}
\toprule
& \multicolumn{6}{c}{LPIPS*1000 $\downarrow$} & \multicolumn{6}{c}{PSNR $\uparrow$} & \multicolumn{6}{c}{SSIM $\uparrow$} \\
\cmidrule(lr){2-7} \cmidrule(lr){8-13} \cmidrule(lr){14-19} 
& \multicolumn{6}{c}{Sequence ID} & \multicolumn{6}{c}{Sequence ID} & \multicolumn{6}{c}{Sequence ID} \\
Method & 377   & 386   & 387   & 392   & 393   & 394   & 377   & 386   & 387   & 392   & 393   & 394   & 377 & 386 & 387 & 392 & 393 & 394 \\
\cmidrule(r){1-1} \cmidrule(lr){2-7} \cmidrule(lr){8-13} \cmidrule(lr){14-19}
Neural Body (14h)& 43.08 & 48.08 & 57.34 & 50.39 & 57.09 & 54.36 & 28.87 & 30.12 & 26.76 & 29.84 & 27.80 & 28.90 & 0.9609 & 0.9588 & 0.9462 & 0.9575 & 0.9472 & 0.9497 \\[1mm]
HumanNeRF-UW (72h)    & \underline{30.28} & \textbf{34.16} & \textbf{41.10} & \underline{36.24} & \textbf{40.30} & \underline{38.12} & \textbf{30.19} & 32.83 & 28.06 & 30.91 & 28.44 & \textbf{30.47} & \underline{0.9642} & \underline{0.9669} & \textbf{0.9551} & \underline{0.9641} & \textbf{0.9544} & \textbf{0.9567} \\
Ours - reconstruction (2-3h) & \textbf{27.72} & \underline{34.42} & \underline{43.44} & \textbf{35.68} & \underline{41.28} & \textbf{38.00} & \underline{30.00} & \textbf{32.90} & \textbf{28.08} & \textbf{31.08} & \textbf{28.51} & \underline{30.28} & \textbf{0.9710} & \textbf{0.9687} & \underline{0.9545} & \textbf{0.9645} & \underline{0.9535} & \underline{0.9564} \\ \midrule
Ours - real-time (40FPS)  & 32.76 & 35.01 & 42.87 & 37.79 & 41.86 & 40.53 & 28.42 & 32.19 & 27.76 & 30.35 & 27.91 & 29.46 & 0.9671 & 0.9677 & 0.9540 & 0.9635 & 0.9531 & 0.9555 \\
\bottomrule
\end{tabular}
}
}
\vspace{-0.5em}
\caption{\textbf{Results on the ZJU-mocap benchmark.}
Our method obtains the best reconstruction results on most of the scenes on the PSNR metric, and on half of the scenes, with respect to the LPIPS and SSIM metrics, while being significantly faster to train (2--3h vs 14h or 72h). Our real-time variant introduces a small performance drop but mantains similar visual quality (cf. Fig.~\ref{fig:zju_qual}).
}
\label{tab:zju}
\end{table*}

\paragraph{Limitations.}

The proposed local emission-absorption raymarching relies on the rasterization of the posed customized human mesh to define the local ray segments where ray marching is performed. This can cause small issues around occlusion boundaries, if the estimated mesh is either too small or too large. If the occluding part of the object is too small, the render will show a slight ``loss of mass" when rendered. If it is too large, this additional occluding part of the object will produce "black pixels", which should in reality show the background part of the object. 
This issue is minimized by making the scaffold as tight as possible.

\section{Experimental evaluation}%
\label{sec:experimental_evaluation}

In order to demonstrate the performance of our proposed method, we evaluate results on two different benchmarks, the ZJU-Mocap scenes and the NeuMan scenes.
In addition, we show the results of the proposed on-device real-time renderer and ablations on the color model employed by the factorized volumetric representation.

\begin{figure}[ht]
\centering
\includegraphics[width=\columnwidth]{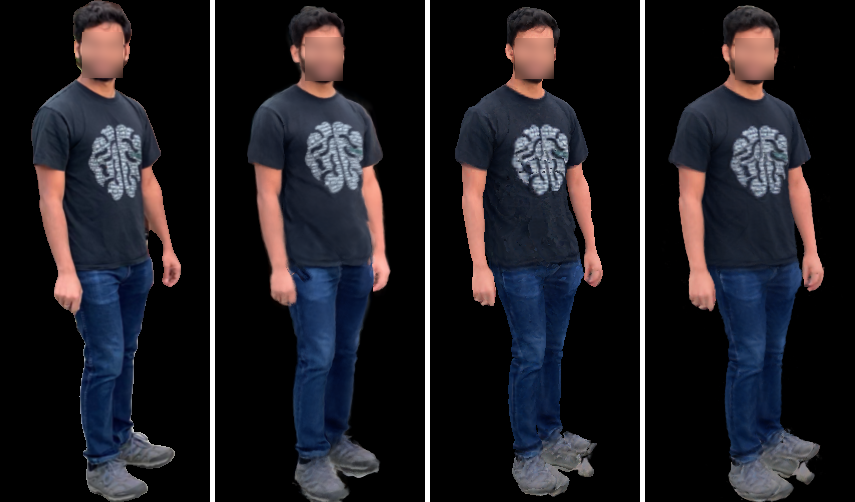}
\includegraphics[width=\columnwidth]{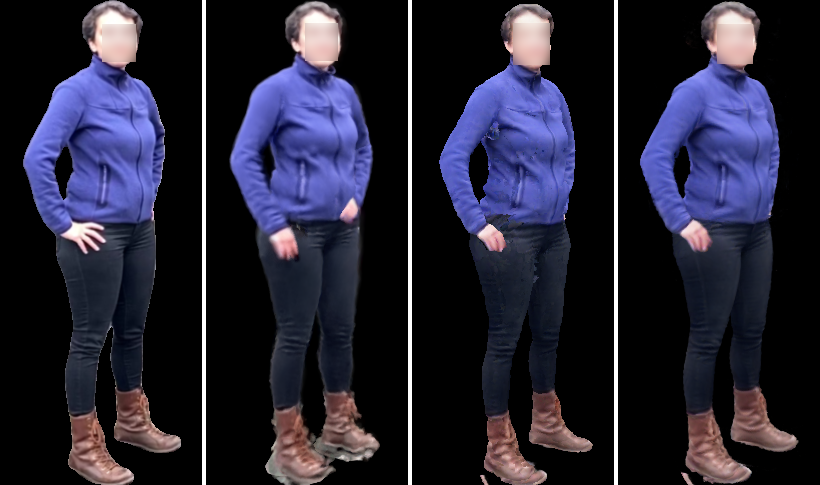}
\includegraphics[width=\columnwidth]{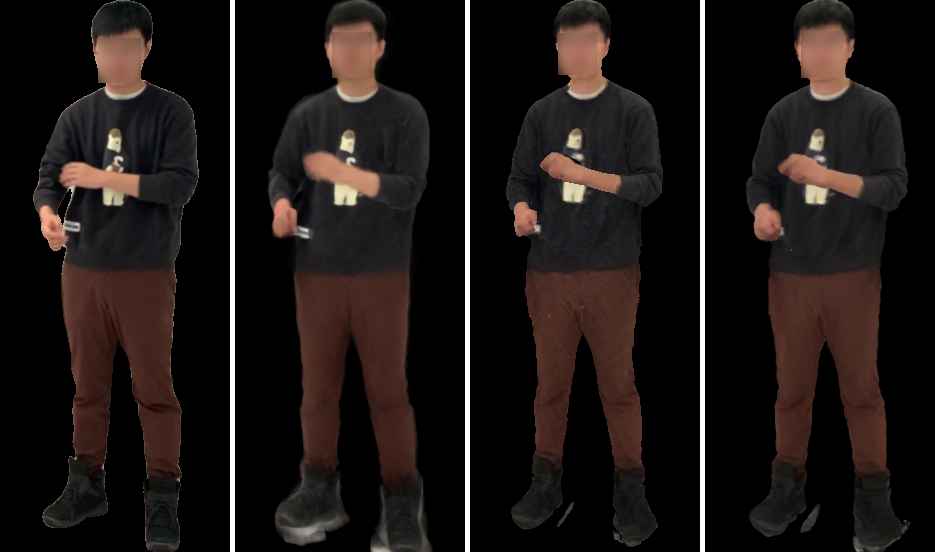}
{\small
{\footnotesize \setlength{\tabcolsep}{0pt}
\begin{tabular}{p{2.1cm}p{2.1cm}p{2.1cm}p{2.1cm}p{2.1cm}}
\centering Ground-truth & 
\centering NeuMan &
\centering Ours - recon. & 
\centering Ours - real-time
\end{tabular}
}
\vspace*{-4mm}
\caption{\textbf{Qualitative results on NeuMan scenes.} 
Our method obtains state-of-the-art visual quality, comparable to that of NeuMan, while being fast to train and render. Faces have been blurred for privacy reasons.}
\label{fig:qual_neuman}\vspace*{-3mm}
}
\end{figure}

\begin{table*}[ht]
\resizebox{1.01\textwidth}{!}{
{\footnotesize
\setlength{\tabcolsep}{1.55pt}
\begin{tabular}{@{}lcccccccccccccccccc@{}}
\toprule
& \multicolumn{6}{c}{LPIPS*1000 $\downarrow$} & \multicolumn{6}{c}{PSNR $\uparrow$} & \multicolumn{6}{c}{SSIM $\uparrow$} \\
\cmidrule(lr){2-7} \cmidrule(lr){8-13} \cmidrule(lr){14-19} 
& \multicolumn{6}{c}{Sequence name} & \multicolumn{6}{c}{Sequence name} & \multicolumn{6}{c}{Sequence name} \\
Method & {\scriptsize bike} & {\scriptsize citron} & {\scriptsize jogging} & {\scriptsize lab} & {\scriptsize parkinglot} & {\scriptsize seattle} & {\scriptsize bike} & {\scriptsize citron} & {\scriptsize jogging} & {\scriptsize lab} & {\scriptsize parkinglot} & {\scriptsize seattle} & {\scriptsize bike} & {\scriptsize citron} & {\scriptsize jogging} & {\scriptsize lab} & {\scriptsize parkinglot} & {\scriptsize seattle} \\
\cmidrule(r){1-1} \cmidrule(lr){2-7} \cmidrule(lr){8-13} \cmidrule(lr){14-19}
NeuMan & \textbf{44.65} & \textbf{28.00} & \textbf{41.53} & 43.38 & \textbf{44.23} & 24.23 & \textbf{26.73} & \textbf{27.88} & \textbf{26.31} & \textbf{28.37} & \textbf{27.43} & 27.80 & \textbf{0.9521} & \textbf{0.9633} & \textbf{0.9496} & \textbf{0.9593} & \textbf{0.9581} & 0.9687 \\
Ours - reconstruction & 49.63 & 29.60 & 42.00 & \textbf{42.18} & 49.24 & \textbf{19.76} & 26.37 & 26.60 & 25.88 & 28.29 & 24.12 & \textbf{28.43} & 0.9465 & 0.9588 & 0.9461 & 0.9574 & 0.9524& 0.\textbf{9722} \\\midrule
Ours - real-time & 48.04 & 31.43 & 41.86 & 42.71 & 56.54 & 21.14 & 25.86 & 26.12 & 25.13 & 27.39 & 23.12 & 27.86 & 0.9457 & 0.9580 & 0.9451 & 0.9552 & 0.9438 & 0.9706 \\
\bottomrule
\end{tabular}
}
}
\vspace{-0.5em}
\caption{\textbf{Quantitative results on the NeuMan benchmark.}
Our method obtains comparable reconstruction performance to NeuMan~\cite{neuman} for most sequences, with superior performance for \emph{seattle}, while having real-time rendering capabilities without significant loss in visual quality (cf. Fig.~\ref{fig:qual_neuman}). 
}%
\label{tab:neuman}
\end{table*}

\paragraph{Implementation details.}
We implement our model using PyTorch~\cite{pytorch}. In particular, we use 8 channels for modeling density and color ($R_\sigma {=} R_c {=} 8$) in the tensorial model. In addition, we use a coarse-to-fine training approach, increasing the voxel grid resolution $D\times H\times W$ several times during training, following~\cite{tensorf}. We begin training with a resolution $HWD=10^6$ and end with $HWD=4.096 \times 10^6$, scaled accordingly to the person's bounding box. At each training iteration, a batch of training rays $r_u$ is constructed by sampling six $32\times 32$ image patches, each centered around a foreground point. The models are trained for 30 epochs of 1000 iterations each, regardless of the number of training images. Before posing the SMPL template with the provided $\theta_{\text{data}}$ parameters from each dataset, we compute a small refinement $\theta_{\text{ref}}$ using a 4-layer MLP with 256 hidden units, following~\cite{humannerf-uw}. The final body parameters are thus $\theta = \theta_{\text{ref}} \circ \theta_{\text{data}}$. The factorized neural field and the pose correction MLP are trained jointly. For training, we employ a photometric loss, a sparsity regularization loss, as well as a perceptual loss (LPIPS). The whole training procedure takes between 2 and 3 hours on a single modern GPU. 

\paragraph{ZJU-Mocap benchmark.}

We first evaluate our method on the ZJU-Mocap dataset introduced in~\cite{neuralbody}.
We follow the evaluation protocol from~\cite{neuralbody} and, for each sequence, use images from 19 unseen cameras for testing, sampled at a 30 frames interval for the first 300 frames, giving 190 test images per sequence.
Following~\cite{humannerf-uw}, we evaluate on 6 sequences, identified by their sequence IDs (377, 386, 387, 392, 393, and 394). In \cref{tab:zju} and Fig.~\ref{fig:zju_qual} we present the quantitative and qualitative results of our method compared to NeuralBody~\cite{neuralbody} and HumanNeRF-UW~\cite{humannerf-uw}.
We present results both for the reconstruction phase (Sec.~\ref{sec:proposed_method}) using standard emission-absorption raymarching and for the proposed real-time rendering (Sec.~\ref{sec:rt}) that can run at 40FPS on mobile hardware. Our reconstructions obtain the best PSNR results for most of the scenes, and best LPIPS and SSIM results for half of the scenes, while being significantly faster to train than the baselines (2--3h for our method vs. 72h for HumanNeRF-UW and 15h for Neural Body). In addition, we observe that the rasterization-guided raymarching introduces a small performance loss, while allowing for real-time rendering. As shown in Fig.~\ref{fig:zju_qual}, the visual quality of our reconstuction and real-time variants is on-par or superior to previous methods.

\paragraph{NeuMan benchmark.}

In addition, we evaluate our method on the scenes from NeuMan~\cite{neuman}. The NeuMan method uses a NeRF~\cite{nerf} module to model the background scene, which is used for segmenting the person.
As our model does not explicitly model the background, we use XMeM~\cite{xmem} to segment the foreground person by scribbling the first frame of each sequence and propagating results automatically.
The results in \cref{tab:neuman} show that our reconstructions outperform NeuMan on two scenes according to LPIPS, and are slightly inferior for the other four scenes.
Nevertheless, qualitative results from \cref{fig:qual_neuman} show that the perceptual quality of our method is similar to that of NeuMan, even for our real-time renders.

\begin{table}[t]
{\footnotesize
\setlength{\tabcolsep}{3.6pt}
\begin{tabular}{@{}lllccc@{}}
\toprule
                   & \begin{tabular}[c]{@{}l@{}}Color\\ model\end{tabular} & \begin{tabular}[c]{@{}l@{}}Pose\\ correction\end{tabular} & \begin{tabular}[c]{@{}l@{}}LPIPS\\ *1000\end{tabular}$\downarrow$ & \multicolumn{1}{c}{PSNR$\uparrow$} & \multicolumn{1}{c}{SSIM$\uparrow$} \\ \midrule
Ours - recon. & Direct & Yes    & 36.76 & 30.14 & 0.9614\\
FA-Spherical Harm. & SH        & Yes    & 36.97 & 30.17 & 0.9612\\
FA-MLP color       & MLP       & Yes    & 37.57 & 30.20 & 0.9616\\
FA-No pose corr.   & Direct    & No     & 38.63	& 29.95 & 0.9607\\ \bottomrule
\end{tabular}
}
\vspace{-0.5em}
\caption{\textbf{Ablation study.} We  analyze the performance drop incurred by modifying several components of the proposed method.
We show the mean LPIPS, PSNR and SSIM metrics over all ZJU-Mocap sequences.}\label{tab:ablation}
\vspace*{-5mm}
\end{table}

\paragraph{Model ablations.}

We also evaluate the effect of modifying the color model of the factorized volumetric representation.
In \cref{sec:radiance_fields} we proposed to model color directly, \ie outputting RGB values directly from the factorized volumetric representation.
Alternatively, we can have the factorized representation output spherical harmonic~\emph{coefficients} or color \emph{descriptors} which are then used to and compute the final RGB values via spherical harmonic functions or a small color MLP, respectively.
The results of using these different color models are presented in \cref{tab:ablation}.
We observe that the choice of the color model does not affect performance significantly.
We therefore propose to use direct color prediction (RGB values), for its simplicity compared to using spherical harmonics or MLP for color prediction, allowing for faster real-time rendering.
Finally, we observe that disabling the body pose correction also diminishes performance to a small extent.

\section{Conclusion}%
\label{sec:conclusions}

We have introduced a method capable of learning neural radiance fields of articulated humans from monocular videos.
By adopting a factorized radiance field model and a simple deformation model, we were able to cut down the training time by an order of magnitude or more compared to prior work, while matching these methods in visual quality.
Furthermore, we have introduced a novel approach to render these dynamic reconstructions in real-time on a mobile GPU. This is done through a rasterization-guided local raymarching which leverages a refined customized human mesh for rendering the neural radiance field without resorting to baking, and with minimal loss of quality. 
As far as we know, this is the first work to perform realtime rendering of dynamic humans in mobile hardware using radiance fields, and without resorting to baking. We hope this work will inspire other future work in this area.

{\small\bibliographystyle{ieee_fullname}\bibliography{shortstrings, bibliography}}

\begin{thebibliography}{10}\itemsep=-1pt

\bibitem{mipnerf}
Jonathan~T Barron, Ben Mildenhall, Matthew Tancik, Peter Hedman, Ricardo
  Martin-Brualla, and Pratul~P Srinivasan.
\newblock Mip-nerf: A multiscale representation for anti-aliasing neural
  radiance fields.
\newblock In {\em Proc. ICCV}, 2021.

\bibitem{mipnerf360}
Jonathan~T. Barron, Ben Mildenhall, Dor Verbin, Pratul~P. Srinivasan, and Peter
  Hedman.
\newblock Mip-nerf 360: Unbounded anti-aliased neural radiance fields.
\newblock In {\em Proc. CVPR}, 2022.

\bibitem{smplify}
Federica Bogo, Angjoo Kanazawa, Christoph Lassner, Peter Gehler, Javier Romero,
  and Michael~J Black.
\newblock Keep it {SMPL}: Automatic estimation of 3d human pose and shape from
  a single image.
\newblock In {\em Proc. ECCV}, 2016.

\bibitem{eggan}
Eric~R Chan, Connor~Z Lin, Matthew~A Chan, Koki Nagano, Boxiao Pan, Shalini
  De~Mello, Orazio Gallo, Leonidas~J Guibas, Jonathan Tremblay, Sameh Khamis,
  et~al.
\newblock Efficient geometry-aware 3d generative adversarial networks.
\newblock In {\em Proc. CVPR}, 2022.

\bibitem{tensorf}
Anpei Chen, Zexiang Xu, Andreas Geiger, Jingyi Yu, and Hao Su.
\newblock {TensoRF}: Tensorial radiance fields.
\newblock In {\em Proc. ECCV}, 2022.

\bibitem{chen21snarf:}
Xu Chen, Yufeng Zheng, Michael~J. Black, Otmar Hilliges, and Andreas Geiger.
\newblock {SNARF:} differentiable forward skinning for animating non-rigid
  neural implicit shapes.
\newblock {\em arXiv.cs}, abs/2104.03953, 2021.

\bibitem{mobilenerf}
Zhiqin Chen, Thomas Funkhouser, Peter Hedman, and Andrea Tagliasacchi.
\newblock Mobilenerf: Exploiting the polygon rasterization pipeline for
  efficient neural field rendering on mobile architectures.
\newblock {\em arXiv preprint arXiv:2208.00277}, 2022.

\bibitem{xmem}
Ho~Kei Cheng and Alexander~G. Schwing.
\newblock {XMem}: Long-term video object segmentation with an atkinson-shiffrin
  memory model.
\newblock In {\em Proc. ECCV}, 2022.

\bibitem{minkowskiengine}
Christopher Choy, JunYoung Gwak, and Silvio Savarese.
\newblock 4d spatio-temporal convnets: Minkowski convolutional neural networks.
\newblock In {\em Proc. CVPR}, 2019.

\bibitem{deng20nasa}
Boyang Deng, John~P. Lewis, Timothy Jeruzalski, Gerard Pons{-}Moll, Geoffrey~E.
  Hinton, Mohammad Norouzi, and Andrea Tagliasacchi.
\newblock {NASA}: Neural articulated shape approximation.
\newblock In {\em Proc. ECCV}, 2020.

\bibitem{fastnerf}
Stephan~J Garbin, Marek Kowalski, Matthew Johnson, Jamie Shotton, and Julien
  Valentin.
\newblock {FastNeRF}: High-fidelity neural rendering at 200fps.
\newblock In {\em Proc. ICCV}, 2021.

\bibitem{surfacesimplif}
Michael Garland and Paul~S Heckbert.
\newblock Surface simplification using quadric error metrics.
\newblock In {\em Proceedings of the 24th annual conference on Computer
  graphics and interactive techniques}, 1997.

\bibitem{sparseconvnet}
Benjamin Graham and Laurens van~der Maaten.
\newblock Submanifold sparse convolutional networks.
\newblock {\em arXiv preprint arXiv:1706.01307}, 2017.

\bibitem{archpp}
Tong He, Yuanlu Xu, Shunsuke Saito, Stefano Soatto, and Tony Tung.
\newblock {ARCH++}: Animation-ready clothed human reconstruction revisited.
\newblock In {\em Proc. ICCV}, 2021.

\bibitem{snerg}
Peter Hedman, Pratul~P Srinivasan, Ben Mildenhall, Jonathan~T Barron, and Paul
  Debevec.
\newblock Baking neural radiance fields for real-time view synthesis.
\newblock In {\em Proc. ICCV}, 2021.

\bibitem{arch}
Zeng Huang, Yuanlu Xu, Christoph Lassner, Hao Li, and Tony Tung.
\newblock {ARCH}: Animatable reconstruction of clothed humans.
\newblock In {\em Proc. CVPR}, 2020.

\bibitem{neuman}
Wei Jiang, Kwang~Moo Yi, Golnoosh Samei, Oncel Tuzel, and Anurag Ranjan.
\newblock {NeuMan}: Neural human radiance field from a single video.
\newblock In {\em Proc. ECCV}, 2022.

\bibitem{screenedpoission}
Michael Kazhdan and Hugues Hoppe.
\newblock Screened poisson surface reconstruction.
\newblock {\em ACM Transactions on Graphics}, 32(3), 2013.

\bibitem{spin}
Nikos Kolotouros, Georgios Pavlakos, Michael~J Black, and Kostas Daniilidis.
\newblock Learning to reconstruct 3d human pose and shape via model-fitting in
  the loop.
\newblock In {\em Proc. ICCV}, 2019.

\bibitem{tava}
Ruilong Li, Julian Tanke, Minh Vo, Michael Zollhofer, Jurgen Gall, Angjoo
  Kanazawa, and Christoph Lassner.
\newblock {TAVA}: Template-free animatable volumetric actors.
\newblock In {\em Proc. ECCV}, 2022.

\bibitem{nsff}
Zhengqi Li, Simon Niklaus, Noah Snavely, and Oliver Wang.
\newblock Neural scene flow fields for space-time view synthesis of dynamic
  scenes.
\newblock In {\em Proc. CVPR}, 2021.

\bibitem{neuralactor}
Lingjie Liu, Marc Habermann, Viktor Rudnev, Kripasindhu Sarkar, Jiatao Gu, and
  Christian Theobalt.
\newblock Neural actor: Neural free-view synthesis of human actors with pose
  control.
\newblock {\em ACM Transactions on Graphics}, 40(6), 2021.

\bibitem{smpl}
Matthew Loper, Naureen Mahmood, Javier Romero, Gerard Pons-Moll, and Michael~J
  Black.
\newblock {SMPL}: A skinned multi-person linear model.
\newblock {\em ACM Transactions on Graphics}, 34(6), 2015.

\bibitem{nerf}
Ben Mildenhall, Pratul~P Srinivasan, Matthew Tancik, Jonathan~T Barron, Ravi
  Ramamoorthi, and Ren Ng.
\newblock {NeRF}: Representing scenes as neural radiance fields for view
  synthesis.
\newblock {\em Communications of the ACM}, 65(1), 2022.

\bibitem{nguyen2021human}
Phong Nguyen, Nikolaos Sarafianos, Christoph Lassner, Janne Heikkila, and Tony
  Tung.
\newblock Free-viewpoint {RGB-D} human performance capture and rendering.
\newblock In {\em Proc. ECCV}, 2022.

\bibitem{nerfies}
Keunhong Park, Utkarsh Sinha, Jonathan~T Barron, Sofien Bouaziz, Dan~B Goldman,
  Steven~M Seitz, and Ricardo Martin-Brualla.
\newblock Nerfies: Deformable neural radiance fields.
\newblock In {\em Proc. ICCV}, 2021.

\bibitem{hypernerf}
Keunhong Park, Utkarsh Sinha, Peter Hedman, Jonathan~T. Barron, Sofien Bouaziz,
  Dan~B Goldman, Ricardo Martin-Brualla, and Steven~M. Seitz.
\newblock {HyperNeRF}: A higher-dimensional representation for topologically
  varying neural radiance fields.
\newblock {\em ACM Transactions on Graphics}, 40(6), 2021.

\bibitem{pytorch}
Adam Paszke, Sam Gross, Francisco Massa, Adam Lerer, James Bradbury, Gregory
  Chanan, Trevor Killeen, Zeming Lin, Natalia Gimelshein, Luca Antiga, et~al.
\newblock Pytorch: an imperative style, high-performance deep learning library.
\newblock In {\em NeurIPS}, 2019.

\bibitem{neuralbody}
Sida Peng, Yuanqing Zhang, Yinghao Xu, Qianqian Wang, Qing Shuai, Hujun Bao,
  and Xiaowei Zhou.
\newblock Neural body: Implicit neural representations with structured latent
  codes for novel view synthesis of dynamic humans.
\newblock In {\em Proc. CVPR}, 2021.

\bibitem{dnerf}
Albert Pumarola, Enric Corona, Gerard Pons-Moll, and Francesc Moreno-Noguer.
\newblock {D-NeRF}: Neural radiance fields for dynamic scenes.
\newblock In {\em Proc. CVPR}, 2021.

\bibitem{FVS}
Gernot Riegler and Vladlen Koltun.
\newblock Free view synthesis.
\newblock In {\em Proc. ECCV}, 2020.

\bibitem{SVS}
Gernot Riegler and Vladlen Koltun.
\newblock Stable view synthesis.
\newblock In {\em Proc. CVPR}, 2021.

\bibitem{pifu}
Shunsuke Saito, Zeng Huang, Ryota Natsume, Shigeo Morishima, Angjoo Kanazawa,
  and Hao Li.
\newblock {PIFu}: Pixel-aligned implicit function for high-resolution clothed
  human digitization.
\newblock In {\em Proc. ICCV}, 2019.

\bibitem{pifuhd}
Shunsuke Saito, Tomas Simon, Jason Saragih, and Hanbyul Joo.
\newblock {PIFuHD}: Multi-level pixel-aligned implicit function for
  high-resolution 3d human digitization.
\newblock In {\em Proc. CVPR}, 2020.

\bibitem{vggnet}
Karen Simonyan and Andrew Zisserman.
\newblock Very deep convolutional networks for large-scale image recognition.
\newblock In {\em Proc. ICLR}, 2015.

\bibitem{anerf}
Shih-Yang Su, Frank Yu, Michael Zollhoefer, and Helge Rhodin.
\newblock {A-NeRF}: Surface-free human 3d pose refinement via neural rendering.
\newblock In {\em NeurIPS}, 2021.

\bibitem{romp}
Yu Sun, Qian Bao, Wu Liu, Yili Fu, Michael~J Black, and Tao Mei.
\newblock Monocular, one-stage, regression of multiple 3d people.
\newblock In {\em Proc. CVPR}, 2021.

\bibitem{neus}
Peng Wang, Lingjie Liu, Yuan Liu, Christian Theobalt, Taku Komura, and Wenping
  Wang.
\newblock Neus: Learning neural implicit surfaces by volume rendering for
  multi-view reconstruction.
\newblock In {\em NeurIPS}, 2021.

\bibitem{ibrnet}
Qianqian Wang, Zhicheng Wang, Kyle Genova, Pratul~P Srinivasan, Howard Zhou,
  Jonathan~T Barron, Ricardo Martin-Brualla, Noah Snavely, and Thomas
  Funkhouser.
\newblock {IBRNet}: Learning multi-view image-based rendering.
\newblock In {\em Proc. CVPR}, 2021.

\bibitem{vid2actor}
Chung-Yi Weng, Brian Curless, and Ira Kemelmacher-Shlizerman.
\newblock {Vid2Actor}: Free-viewpoint animatable person synthesis from video in
  the wild.
\newblock {\em arXiv preprint arXiv:2012.12884}, 2020.

\bibitem{humannerf-uw}
Chung-Yi Weng, Brian Curless, Pratul~P Srinivasan, Jonathan~T Barron, and Ira
  Kemelmacher-Shlizerman.
\newblock {HumanNeRF}: Free-viewpoint rendering of moving people from monocular
  video.
\newblock In {\em Proc. CVPR}, 2022.

\bibitem{plenoctrees}
Alex Yu, Ruilong Li, Matthew Tancik, Hao Li, Ren Ng, and Angjoo Kanazawa.
\newblock {PlenOctrees} for real-time rendering of neural radiance fields.
\newblock In {\em Proc. ICCV}, 2021.

\bibitem{lpips}
Richard Zhang, Phillip Isola, Alexei~A Efros, Eli Shechtman, and Oliver Wang.
\newblock The unreasonable effectiveness of deep features as a perceptual
  metric.
\newblock In {\em Proc. CVPR}, 2018.

\bibitem{humannerf-shanghai}
Fuqiang Zhao, Wei Yang, Jiakai Zhang, Pei Lin, Yingliang Zhang, Jingyi Yu, and
  Lan Xu.
\newblock {HumanNeRF}: Efficiently generated human radiance field from sparse
  inputs.
\newblock In {\em Proc. CVPR}, 2022.

\end{thebibliography}
\clearpage
\begin{appendices}

\section{Additional technical details}

\subsection{Linear blend skinning\label{appendix:lbs}}

Given a canonical mesh and its associated \emph{skeleton}, which is a collection of bones $b\in\{1,\dots,B\}$, and the pose parameters $\theta$, which are the rotations of the bone joints, linear blend skinning (LBS) allows to~\emph{pose} the canonical mesh using the pose parameters alone. In order to account for transitions between bones, each canonical vertex $\bar{\bx}$ is not assigned to a single bone, but rather~\emph{softly} attached to different bones by its \emph{skinning weights} $w(\bar{\bx})$.

Bones form a tree, where $b'=\operatorname{par}(b)$ denotes the parent of bone $b$, except for the root bone $b=1$ which has no parent.

Then, given a 3D point $\tilde{\bx}$ defined in the reference frame of bone $b$, we can express it in the reference frame of the parent bone as $\bx' = g_b(\theta) \tilde{\bx}$, where $g_b(\theta)\in SE(3)$ is the rigid transformation between the two bones.
Hence, the pose vector $\theta$ specifies three rotation angles for each transformation $g_b$ (the translation component is given by the parent bone's length); only for the root transformation $g_1$, which positions the skeleton in world space, $\theta$ specifies the translation component as well.
In order to express the 3D point in world coordinates $\bx = G_b(\theta) \tilde{\bx}$, we compose recursively the transformation towards the root:
$$
G_b(\theta) = G_{\operatorname{par}(b)}(\theta) \circ g_b(\theta),~~~G_1(\theta) = g_1(\theta).
$$

Consider now a 3D point $\bar{\bx}$ in canonical space, rigidly attached to bone $b$.
In order to pose it, we first express it relative to its bone as $\tilde{\bx} = G_b^{-1}(\theta_0)\bar{\bx}$, where $\theta_0$ is the \emph{canonical pose}.
Then, we map it to world space by as $\bx = G_b(\theta)\tilde{\bx}$.
In practice, each point $\bar{\bx}$ is attached to all bones to a degree expressed by the \emph{skinning weights} $w(\bar{\bx}) \in \mathbb{R}^B$, which are non-negative and sum to one.
The individual maps are linearly combined (called \emph{blend skinning}), resulting in the affine map
\begin{equation}\label{eq:skinning}
A(\bar{\bx}; \theta)
=
\sum_{b=1}^B
w_b(\bar{\bx})~
G_b(\theta) \circ G_b^{-1}(\theta_0).
\end{equation}
For simplicity, in the main paper we consider the composed affine map for each bone  $A_b(\theta) = G_b(\theta) \circ G_b^{-1}(\theta_0)$, assuming $\theta_0$ as fixed.

Finally, the corresponding posing function resulting from linear blend skinning is
\begin{equation}\label{eq:fwd}
  x = h(\bar{\bx}; \theta) = A(\bar{\bx} ; \theta) \bar{\bx}.
\end{equation}

\subsection{Customized human mesh extraction\label{appendix:mesh}}

While previous methods have used algorithms such as marching cubes to extract meshes from implicit volumetric representations, we propose to use a render-based approach instead, as our canonical model contains noise in the regions where no supervision was provided due to the threshold $\tau$. To this end, we render frames from the trained canonical factorized model, along with their depth maps and estimated foreground masks, with a circular camera trajectory around the object (in a turntable fashion). 

The depth maps are computed by measuring the depth w.r.t.\ to the current camera of the expected termination $\hat{\textbf{x}}_t$ of each camera ray, defined as 
\begin{equation}
    \hat{\textbf{x}}_t =\sum_{i=0,\dots,N-1} (T_i - T_{i+1}) \textbf{x}_i,
\end{equation}
as done in {NeRF~[21]}. Similarly, the foreground masks are obtained by computing the expected ray opacity as
\begin{equation}
\hat{o} =\sum_{i=0,\dots,N-1} (T_i - T_{i+1})= 1 - T_N,    
\end{equation}
The estimated object foreground masks are obtained by retaining the pixels where $\hat{o} > 0.5$. 

Then, each depth-map is \emph{unprojected} using the camera parameters and combined with the color render to obtain a set of 3D points, which are then fused with those from other images to obtain a dense point cloud. This point cloud is finally filtered using the foreground masks, such that only points that are classified as foreground for all images are kept. 

Next, this dense point cloud is converted into a mesh by applying the screened Poisson surface reconstruction algorithm~\cite{screenedpoission}, and simplified using an edge-collapse technique~\cite{surfacesimplif} such that the number of final faces is around ~15K, which is on the order of magnitude to the template mesh in SMPL. The surface normals for the extracted canonical mesh are mapped to the surface normals of the closest vertices of the underlying SMPL template mesh. Note that these normals are not used for shading, but only for running the screened Poisson surface reconstruction.

\begin{figure}[h!]
    \centering
    {\footnotesize
        \begin{tabular}{ccc}
        \includegraphics[width=0.24\columnwidth]{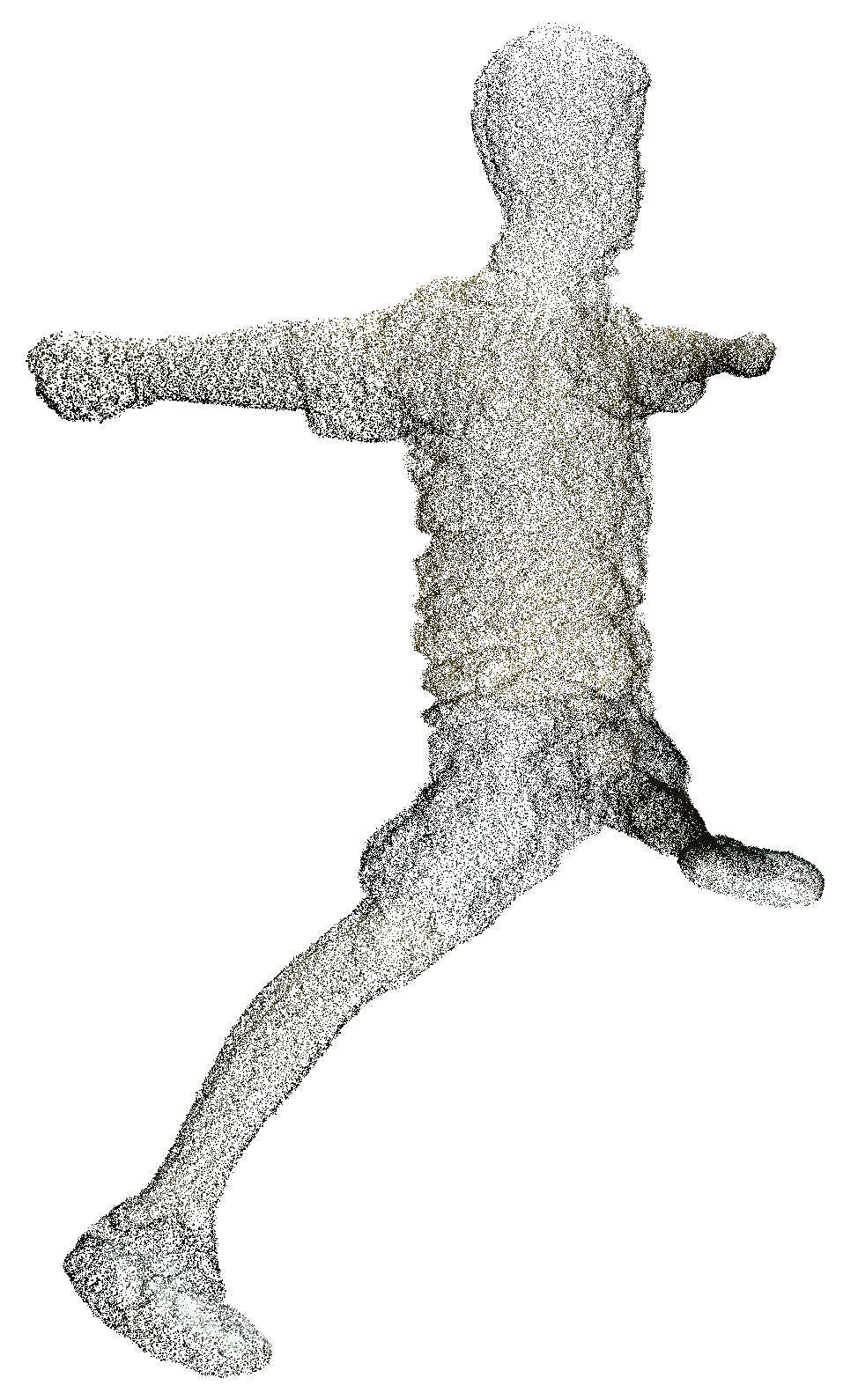} & \includegraphics[width=0.24\columnwidth]{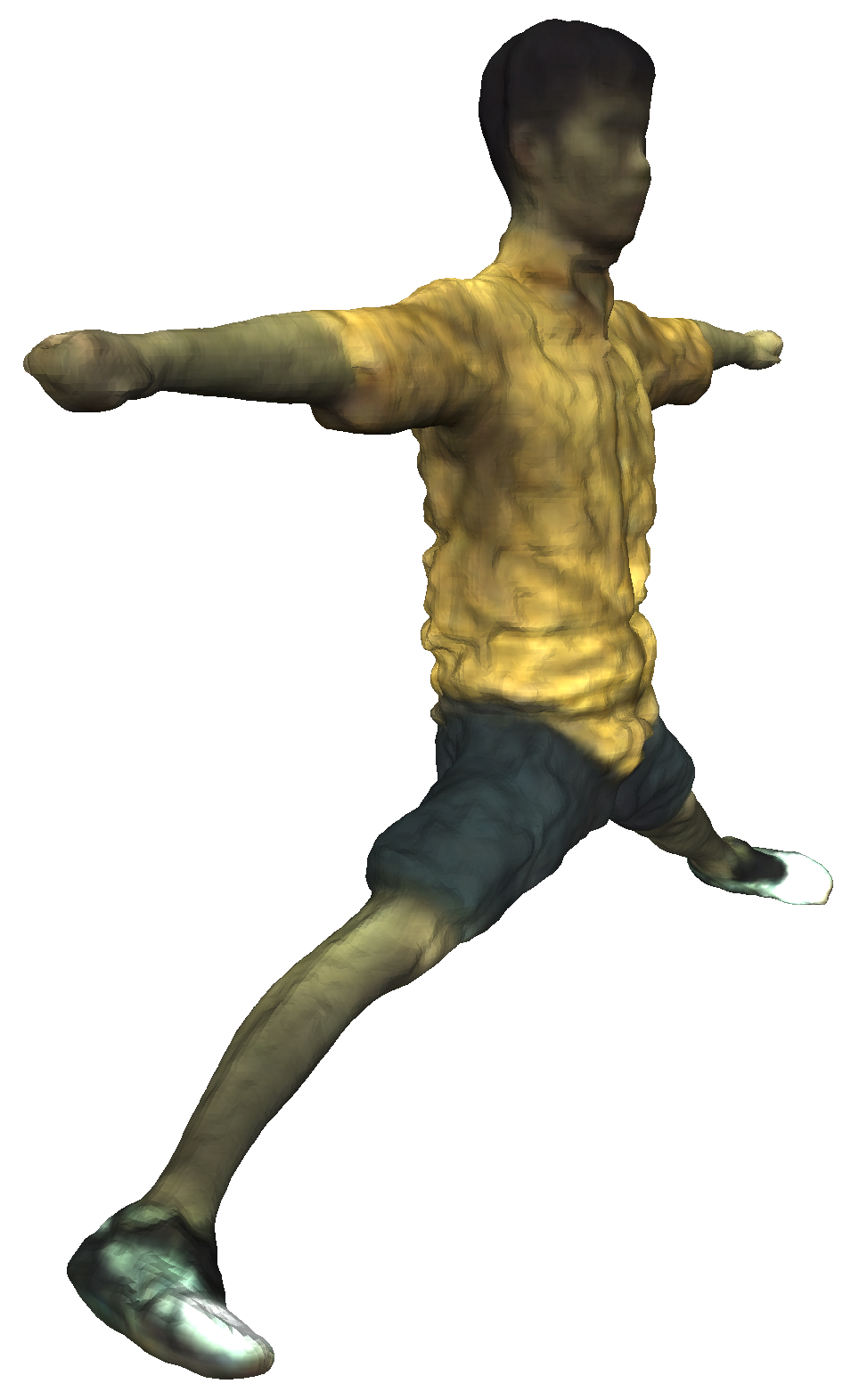} & \includegraphics[width=0.24\columnwidth]{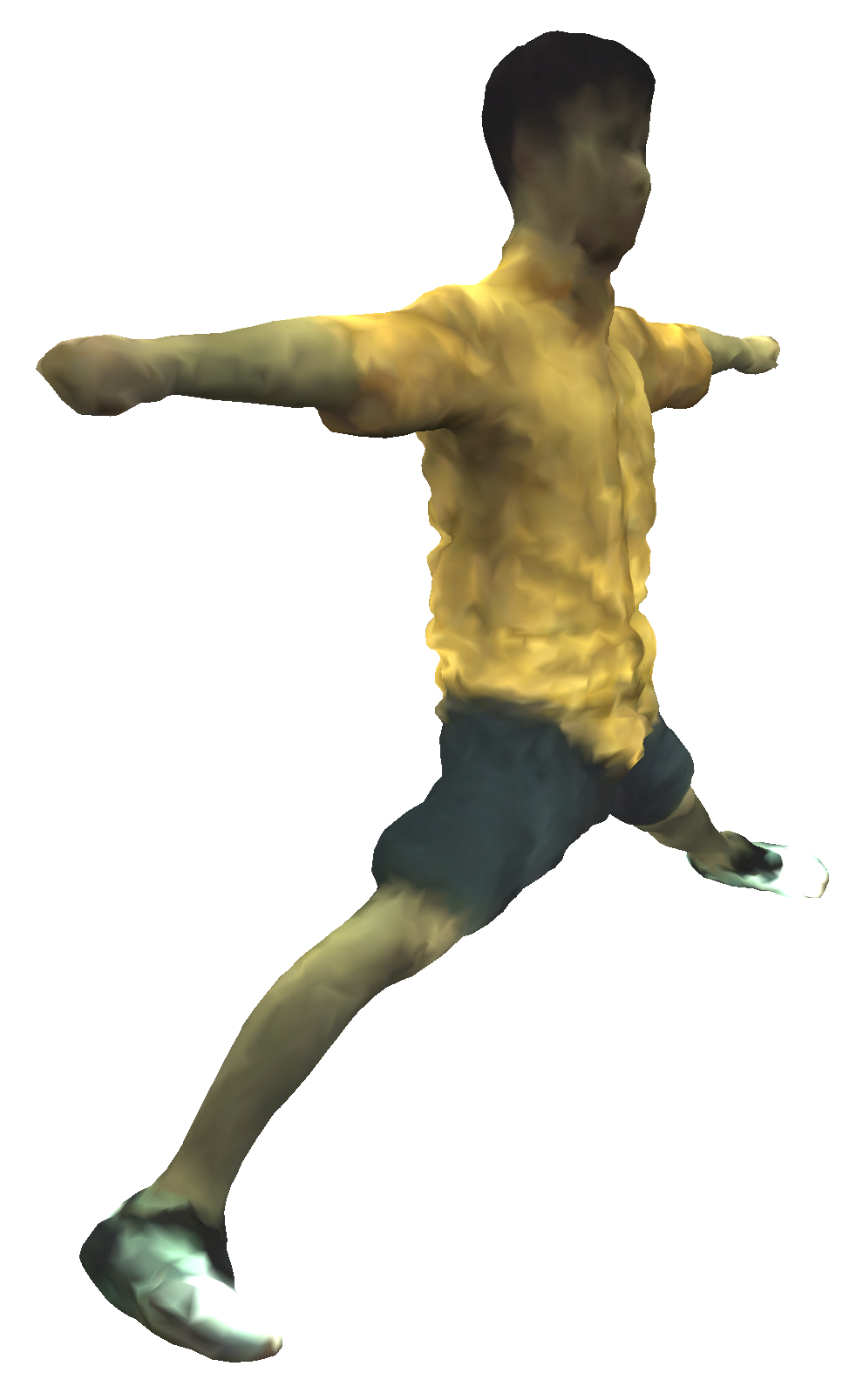} \\
        (a) Dense PCL & (b) Mesh (100K faces) & (c) Mesh (15K faces)
    \end{tabular}
    }
    \caption{\textbf{Canonical mesh extraction.} We first generate a dense point cloud (a), from which we perform screened Poisson reconstruction (b), and, finally mesh simplification to 15K faces (c).}
    \label{fig:mesh_extraction}
\end{figure}

Finally, in order to be able to perform skinning of the extracted canonical mesh, we transfer the skinning weights from the SMPL template using the nearest vertices. The process of canonical mesh extraction is illustrated in Fig.~\ref{fig:mesh_extraction}. Note that the colors of the extracted canonical  mesh are solely used for visualization purposes, and not used in the proposed rasterization-based raymarching, which only requires the vertices and triangles of the canonical mesh to guide the raymarching on the factorized volumetric representation.

\subsection{Training losses}

We present some additional details about the losses used for training. At each training iteration, a batch $\mathcal{B}=\{\mathcal{P}_i\}_{i=1,\dots,6}$ of six image patches $\mathcal{P}$ of size $32\times 32$ is constructed, and the our proposed model is used to estimate the image color $I(u)$ for each pixel $u\in P$. Then we use three different loss terms for optimizing the proposed model: (i) a photometric loss $\mathcal{L}_{\text{rgb}}$, (ii)  a perceptual loss (LPIPS) $\mathcal{L}_{\text{lpips}}$, and, (iii) a sparsity regularization loss $\mathcal{L}_{\text{sparse}}$. Our overall loss is thus:
$$
\mathcal{L}(\mathcal{B}) = \alpha \mathcal{L}_{\text{rgb}}(\mathcal{B}) + \beta \mathcal{L}_{\text{lpips}}(\mathcal{B}) + \gamma \mathcal{L}_{\text{sparse}},
$$
with:
\begin{align}
    \mathcal{L}_{\text{rgb}}(\mathcal{B})&= \frac{1}{N} \sum_{u\in \mathcal{B}} \lVert  I(u) - I_{\text{gt}}(u) \rVert_2^2\\
    \mathcal{L}_{\text{lpips}}(\mathcal{B})&= \frac{1}{6} \sum_{\mathcal{P}
\in\mathcal{B}} \text{LPIPS}_{\text{vgg}}(I(\mathcal{P}),I_{\text{gt}}(\mathcal{P}))\\
    \mathcal{L}_{\text{sparse}}&= \frac{1}{R_\sigma\cdot HWD} \sum_{(M,v) \in \mathcal{W}} \sum_{rxyz} (Mv)^+,
\end{align}
where $N=6\times 32\times 32$ is the total number of pixels in the training batch, $\text{LPIPS}_{\text{vgg}}$ is the perceptual loss proposed in~\cite{lpips} using the VGG-Net feature extractor~\cite{vggnet}, and the sparsity loss is applied over all three factors $\mathcal{W} = \{
(M^{YX}, v^Z), (M^{YZ}, v^X), (M^{XZ}, v^Y)\}$ of the implicit volumetric representation of the density $\sigma$, and across all channels $R_\sigma$.
In addition, the coefficients $\alpha, \beta, \gamma$ which modulate each loss term vary as the training progresses, and we can therefore define them as functions of the training iteration $i=1,\dots,30\,000$:
\begin{align}
    \alpha(i)& = \begin{cases}
         1 - 0.8 \cdot i/10\,000 &  \text{if } i < 10\,000 \\
         0.2 & \text{otherwise} 
    \end{cases}\\ 
    \beta(i)& = \begin{cases}
         0.8 \cdot i / 10\,000 &  \text{if } i < 10\,000 \\
         0.8 & \text{otherwise} 
    \end{cases}\\
    \gamma(i)& =\begin{cases}
        0 & \text{if } i < 2\,000 \\
        8\cdot 10^{-5} & \text{if } 2\,000 \leq i < 4\,000 \\
        5\cdot 10^{-5} & \text{otherwise}
    \end{cases}
\end{align}

\subsection{Proposed rasterization-guided local raymarching vs. phong shading}

\begin{figure}[h!]
    \centering
    \begin{subfigure}{\columnwidth}
    \includegraphics[width=\columnwidth]{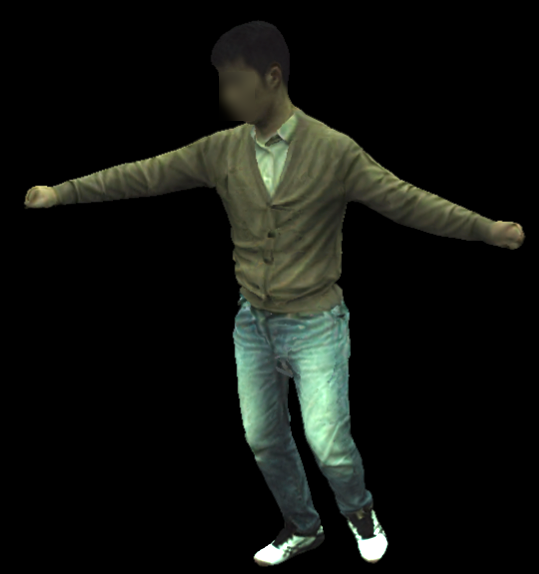}
    \caption{Local volumetric rendering}
    \end{subfigure}
    
    \centering
    \begin{subfigure}{\columnwidth}
    \includegraphics[width=\columnwidth]{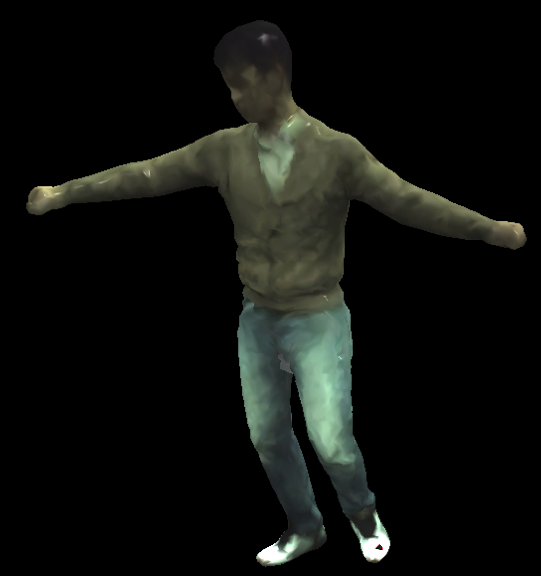}
    \caption{Phong shading}
    \end{subfigure}
    \caption{\textbf{Comparison between rendering techniques.} We illustrate the results of the proposed local rasterization-based emission-absorption raymarching against rendering the template mesh directly with a standard phong shading.}
    \label{fig:phong}
\end{figure}

As an ablation study, we show in~\cref{fig:phong} the comparison of rendering the posed template mesh with our proposed method against using standard phong shading based on vertex colors. Note that as a texture of the human is not available, many details are lost when applying phong shading directly. In addition, this rendering technique requires an artificial light source and generates unrealistic reflection effects. On the contrary, our proposed method recovers the textures details from the learnt volumetric representation, and new views can be rendered directly without any additional artificial light source.

\section{Demos and source code}
\subsection{VR Mixed reality demo}

We have built a VR Mixed reality demo that demonstrates our proposed method, and rendering at 40FPS in consumer-level standalone VR headset, using the reconstructions obtained from the ZJU-mocap scenes.

\begin{figure}[h]
    \centering
    \includegraphics[width=\columnwidth]{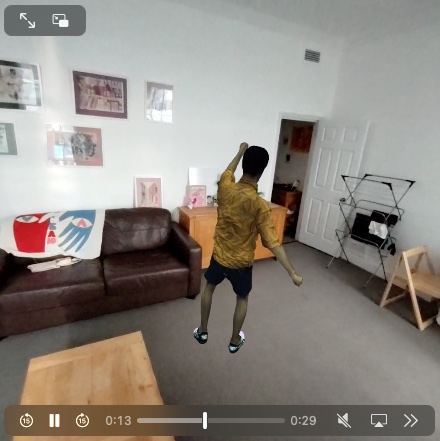}
    \caption{\textbf{VR Mixed reality demo.} The dynamic human is rendered using VR-passthrough at 40FPS on a standalone VR device.}
    \label{fig:vr_demo}
\end{figure}

Several recordings of these scenes playing on the VR headset, as illustrated in \Cref{fig:vr_demo}, are available in our website \url{https://real-time-humans.github.io/}. Note that the background scene is rendered in mixed reality thorough RGB passthrough using the VR headset's frontal RGB camera, giving the effect of an AR device.

\subsection{Desktop WebGL demo}

\begin{figure}[h!]
    \centering
    \includegraphics[width=\columnwidth]{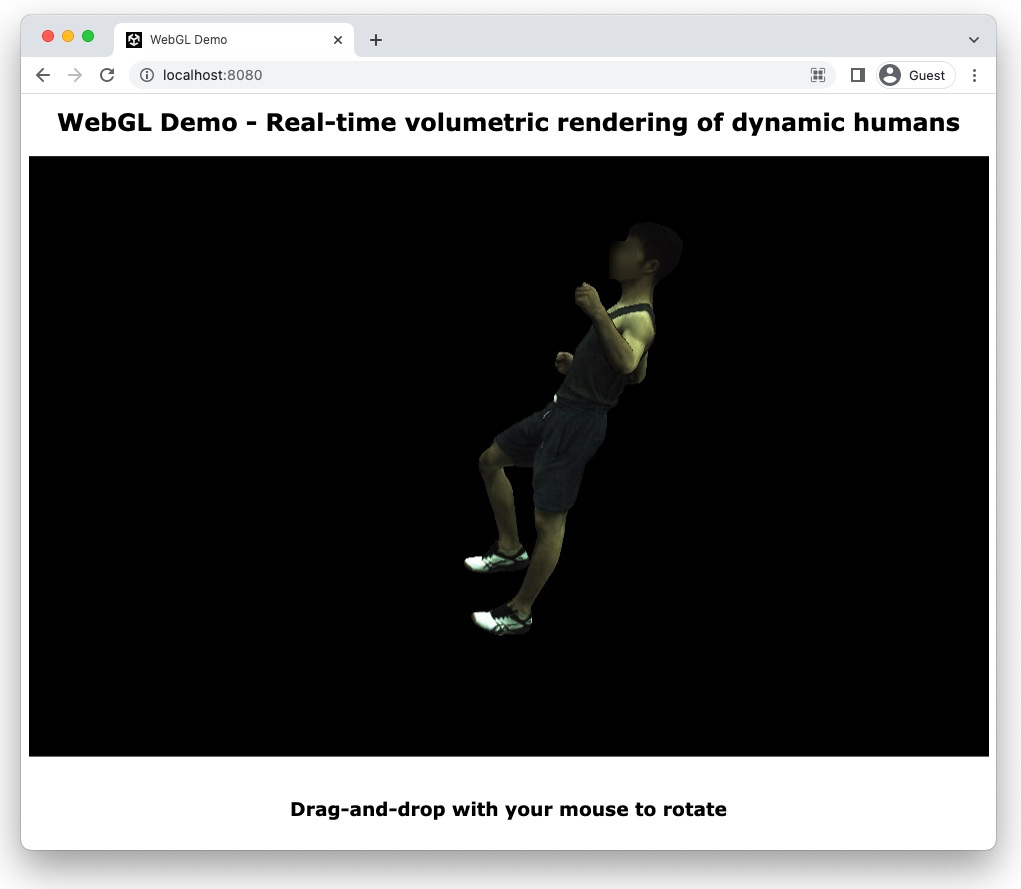}
    \caption{\textbf{WebGL desktop demo.} The demo is available in our website.}
    \label{fig:web_demo}
\end{figure}

Additionally we developed a Desktop WebGL demo that implements the proposed real-time volumetric rendering and runs inside the web browser, as illustrated in \Cref{fig:web_demo}. The demo is available in our website \url{https://real-time-humans.github.io/}.

\subsection{Source code.} We will release our source code to allow for reproducibility. It will be released under an open-source licence. The VR and WebGL demo apps will be released in binary format.

\end{appendices}

\end{document}